\documentclass[runningheads]{llncs}

\usepackage{eccv}

\usepackage{eccvabbrv}

\usepackage{graphicx}
\usepackage{booktabs}

\usepackage{float}
\usepackage{booktabs}      
\usepackage{multirow}      
\usepackage{arydshln}
\usepackage{xcolor}
\usepackage[table]{xcolor}
\usepackage{pifont}
\newcommand{\cmark}{\ding{51}}
\newcommand{\gxmark}{\ding{55}}
\usepackage{caption}
\usepackage{wrapfig}
\usepackage{array}   
\usepackage{makecell}
\newcommand{\ra}[1]{\renewcommand{\arraystretch}{#1}}

\usepackage[accsupp]{axessibility}  % Improves PDF readability for those with disabilities.
\usepackage{marvosym} % for \Letter
\newcommand{\corrauth}{\textsuperscript{\Letter}}

\usepackage{hyperref}

\usepackage{orcidlink}

\begin{document}

\title{Layout-Conditioned Autoregressive Text-to-Image Generation via Structured Masking}
\titlerunning{SMARLI}

\author{ 
Zirui Zheng\inst{1,2} 
\and Takashi Isobe\inst{1}\corrauth 
\and Tong Shen\inst{1} 
\and Xu Jia\inst{2}\corrauth 
\and Jianbin Zhao\inst{2} 
\and Xiaomin Li\inst{1,2} 
\and Mengmeng Ge\inst{1} 
\and Baolu Li\inst{2} 
\and Qinghe Wang\inst{2} 
\and Haiwen Diao\inst{3} 
\and Dong Li\inst{1} 
\and Dong Zhou\inst{1} 
\and Yunzhi Zhuge\inst{2} 
\and Huchuan Lu\inst{2} 
\and Emad Barsoum\inst{1} } 

\authorrunning{Z.~Zheng et al.}

\institute{ 
Advanced Micro Devices Inc. 
\and Dalian University of Technology 
\and S-Lab, Nanyang Technological University }

% \maketitle

% \centering
% \includegraphics[width=1.0\textwidth]{figures/intro/teaser.pdf}
% \captionsetup{type=figure}
% \caption{Example results of layout-conditioned image generation produced by SMARLI. We show that SMARLI achieves fine-grained controllable generation with accurate spatial control and precise rendering of complex attributes, including color, shape, and texture. Global text prompts are omitted for brevity.}
% \label{fig:teaser}

{
    \renewcommand\twocolumn[1][]{#1}
    \maketitle
    \centering
    % \vspace{-6.0mm}
    \includegraphics[width=1.0\textwidth]{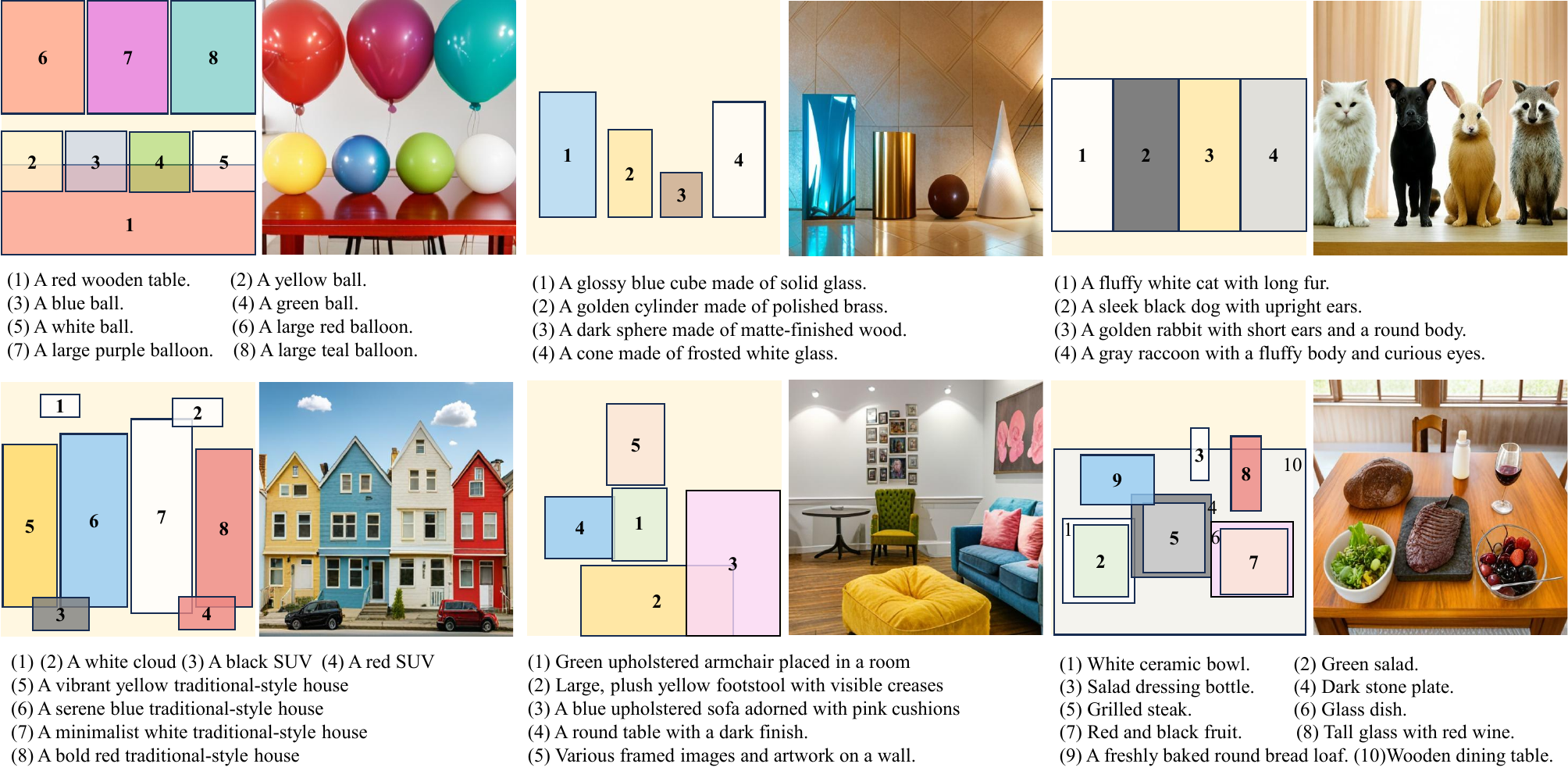}
    \captionsetup{type=figure}
    \vspace{-4.3mm}
    \caption{Example results of layout-conditioned image generation produced by SMARLI. We show that SMARLI achieves fine-grained controllable generation with accurate spatial control and precise rendering of complex attributes, including color, shape, and texture. Global text prompts are omitted for brevity.}
    \label{fig:teaser}
    % \vspace{3.0mm}
}

\begingroup \renewcommand{\thefootnote}{\Letter} \footnotetext{Corresponding authors.} \endgroup

\begin{abstract}
Although autoregressive (AR) models have demonstrated remarkable success in image generation, extending these models to layout-conditioned generation remains challenging due to the sparse nature of layout conditions and the risk of feature entanglement. We present \textbf{S}tructured \textbf{M}asking for \textbf{AR}-based \textbf{L}ayout-to-\textbf{I}mage (SMARLI), a novel framework that effectively integrates spatial layout constraints into the AR generation process. To equip AR models with layout control, a structured masking strategy is applied to the attention computation to govern the interaction among the global prompt, layout, and image tokens. This design prevents the misassociation of different regions with their corresponding descriptions while enabling the sufficient injection of layout constraints into the generation process. To alleviate the exposure bias of AR models and further enhance generation quality and layout accuracy, we incorporate a Group Relative Policy Optimization (GRPO) post-training scheme. We adapt it to the next-set-based paradigm and introduce a specifically designed layout reward, which is coordinated with an image quality reward to guide policy optimization in a balanced manner. Experimental results demonstrate that SMARLI seamlessly integrates layout tokens with text and image tokens without compromising generation quality, and the proposed masking strategy and post-training scheme can also be transferred to standard next-token-based AR models. The proposed framework achieves superior layout control while maintaining the structural simplicity and generation efficiency of AR models.
\end{abstract}   
\section{Introduction}
\label{sec:intro}
The success of AR modeling in Large Language Models~\cite{touvron2023llama, yang2025qwen3} has inspired many methods that extend AR principles to image generation tasks~\cite {dall-e,parti, sun2024llamagen, xie2024showo, wang2024emu3, tian2024visual, mumuni2024survey, xin2025lumina, wu2025harmonizing}, showcasing the effectiveness of AR-based generative frameworks in the vision domain. These approaches reformulate image generation as a next-token, next-set, or next-scale prediction task, demonstrating remarkable potential and emerging as strong competitors to diffusion models. In this work, we investigate the potential of AR models in Layout-to-Image (L2I) generation.

The L2I generation conditions on the spatial locations and semantic descriptions of entities; it has been extensively explored in diffusion models based on different architectures, including U-Net~\cite{li2023gligen, lian2023llm-g, zhou2024migc, cheng2024hico, wang2024instancediffusion, feng2024ranni, dahary2024yourself, yang2024mastering} and Multimodal Diffusion Transformers~\cite{lee2024groundit, zhang2024creatilayout, zhang2025eligen, xianginstanceassemble}. Most existing methods for layout control in diffusion models integrate tokenized layout conditions by introducing additional blocks or even branches. However, such practices typically introduce substantial additional parameters and computational overhead. Furthermore, the introduction of additional blocks compromises the simplicity of the AR transformer architecture and conflicts with the emerging perspective of AR models as a unified generative framework. To avoid the structural modifications of the AR transformer, some prior studies~\cite{li2024controlar, yao2024car, li2024controlvar, mao2025varedit, xu2025scalar} on AR-based conditional image generation adopt a feature-space injection paradigm inspired by ControlNet~\cite{zhang2023adding}. These methods incorporate encoded feature maps of visual condition by either adding them to image token embeddings or integrating them through joint decoding. However, they primarily rely on spatially aligned visual conditions that provide dense visual cues like edge maps or depth maps rather than layout. In contrast, layout conditions are inherently sparse and offer limited visual information, which may become further ambiguous when bounding boxes overlap. Moreover, jointly encoding bounding box coordinates and textual descriptions remains challenging, as it requires a dedicated condition encoder capable of modeling both geometric structure and semantic content.

To avoid specialized condition encoders that convert layouts into feature maps, an alternative paradigm concatenates conditional inputs with the text prompt to form a unified sequence, enabling conditional image generation or image editing within AR models~\cite{chen2025context, mu2025editar, xin2025lumina}.
Despite this advancement, the exploration of this paradigm in L2I generation remains relatively limited. PlanGen~\cite{he2025plangen} emerges as an early attempt that adopts this input sequence concatenation approach. By placing the tokenized layout conditions directly into the input sequence, it avoids complex architectural modifications. However, this approach faces two main challenges: (1) \textbf{semantic interference}, standard AR modeling fails to capture the structural characteristics of layout conditions. With a vanilla causal attention mask applied to the entire input sequence, layout tokens from different regions may interfere across regions, and image tokens may attend to irrelevant layout tokens. This results in feature entanglement and degraded layout precision, particularly in complex scenes. (2) \textbf{Limited image quality}, the visual fidelity of PlanGen remains inferior to that of state-of-the-art diffusion-based L2I models. The model tends to generate noticeable artifacts when synthesizing complex scenes. This limitation restricts its practical applicability in layout-conditioned image synthesis.

In this work, we propose SMARLI, an AR-based L2I framework that integrates a structured masking strategy to mitigate \textbf{semantic interference}, along with a layout-aware GRPO-based post-training scheme to improve \textbf{image quality} and layout controllability. Specifically, we first encode the global text prompt, layout condition, and image into a unified token sequence and feed it into the AR transformer. Then, to mitigate the aforementioned \textbf{interference}, we introduce the structured masking strategy, which explicitly regulates the attention computation across different token types: (1) Each object’s layout tokens attend to the global text prompt tokens for broader contextual cues and to their local layout context, while remaining isolated from the layout tokens of other objects to prevent cross-object interference. (2) Image tokens receive guidance from both global text prompt tokens and region-specific layout tokens, ensuring the generation process is controlled by appropriate contexts while avoiding attribute confusion as illustrated in \cref{fig:teaser}. This introduces inductive priors that align with the sequential organization of prompt, layout, and image tokens, facilitating efficient training without requiring architectural modifications to the base AR models.
% GRPO
Finally, to further improve the \textbf{visual fidelity} and layout controllability, we propose a GRPO-based post-training scheme~\cite{shao2024deepseekmath}, and we adapt it to next-set-based AR models. A human preference score is introduced as a reward to enhance visual fidelity, and a comprehensive layout-aware reward is designed to improve layout controllability. Moreover, GRPO-based post-training mitigates exposure bias in AR models caused by the train–test discrepancy, as it trains on model-generated rollouts.

Extensive experiments demonstrate the effectiveness of the proposed framework in AR-based L2I generation. Furthermore, empirical results indicate that the proposed approach is also applicable to standard next-token-based AR models. It performs favorably against state-of-the-art diffusion-based counterparts, showing further potential of AR models.

Our main contributions are summarized as follows.
(1) We introduce SMARLI, a layout-conditioned AR-based text-to-image (T2I) framework, which incorporates a structured masking strategy to inject layout conditions, mitigating region-description misalignment.
(2) We adopt GRPO-based post-training and propose a comprehensive layout reward that, together with an image-quality reward, notably enhances both image fidelity and layout controllability.
(3) Extensive experiments validate the effectiveness and generality of SMARLI across multiple L2I generation benchmarks.

\section{Related Work}
\label{sec:related_work}

\subsection{Autoregressive Image Generation}

AR models have recently achieved image generation performance comparable to diffusion models. Typically, they employ a discrete VQ-VAE tokenizer~\cite{esser2021taming} to convert images into sequences of tokens, which are then modeled using next-token prediction in raster-scan order~\cite{sun2024llamagen, wang2024emu3, chen2025janus}. To enhance both efficiency and output quality, recent approaches have reformulated the generation process. Masked generative models~\cite{chang2022maskgit,chang2023muse,xie2024showo,li2024mar,deng2024nova, wu2025harmonizing} leverage bidirectional context to predict sets of masked tokens in parallel, while still maintaining AR training. \textsc{VAR}~\cite{tian2024visual} introduces residual vector quantization to enable next-scale prediction, further improving image quality. In this paper, we present SMARLI, an AR-based image generation framework that endows AR models with explicit layout control. The proposed framework integrates global text prompts and spatial layout conditions into the autoregressive generation process, enabling precise spatial control without modifying the AR Transformer architecture.

\subsection{Layout-conditioned T2I Generation}

For layout control, existing diffusion-based T2I models often achieve effective layout control \cite{li2023gligen,wang2024instancediffusion, cheng2024hico, chen2024layout-attn,yang2024mastering, ge2024customizing, dahary2024yourself, zhang2025eligen, ge2026ego} by introducing additional attention between the layout and image or directly manipulating the attention map and image latent. For instance, GLIGEN~\cite{li2023gligen} incorporates layout conditions into newly introduced trainable layers through a gated mechanism. InstanceDiffusion~\cite{wang2024instancediffusion} extends this paradigm to instance-level control, supporting flexible layout specifications. SiamLayout~\cite{zhang2024creatilayout} employs a separate branch of the network to process the layout in MMDiT. RPG~\cite{yang2024mastering} independently generates each region and composes them in the image latent space based on a pre-defined coarse layout. InstanceAssemble~\cite{xianginstanceassemble} proposed Assemble-MMDiT to inject the layout conditions into image tokens after global prompt injection. PlanGen~\cite{he2025plangen} proposed a unified model for text-to-layout generation and layout-conditioned image generation.
In this paper, based on the AR T2I model, we propose a structured masking strategy to realize effective layout control under the unified input sequence paradigm without introducing computationally expensive components.

\subsection{Post-training in Visual Content Generation}

The success of reward-based optimization in content generation~\cite{clark2023directly, xu2023imagereward, peng2025omni, li2026portraitgen} has inspired extensive research on feedback-driven learning. Early works applied RL and preference optimization to diffusion models, including DDPO~\cite{black2023ddpo} and Diffusion-DPO~\cite{wallace2024diffdpo}. Recent studies further generalized reward-based optimization across architectures, such as FlowGRPO~\cite{liu2025flowgrpo}, DanceGRPO~\cite{xue2025Dancegrpo}, and ReNeg \cite{li2025reneg}. In next-token-based AR generation, SimpleAR~\cite{wang2025simplear} and T2I-R1~\cite{jiang2025t2i-r1} employed GRPO post-training to improve text-image alignment.

However, GRPO has not been explored for AR-based layout-to-image (L2I) generation. To bridge this gap, we propose a GRPO-based post-training scheme with a specialized layout reward, complemented by an image quality reward, enabling more precise layout control while improving generation quality.
\section{Method}
\label{sec:method}

We build our framework upon Show-o~\cite{xie2024showo}, which follows the next-set prediction paradigm. As shown in \cref{fig:pipeline}, our framework consists of three main components: (1) \textbf{Tokenization}: The global text prompt, layout, and image are tokenized separately, concatenated into a single sequence, and fed into the AR transformer. (2) \textbf{Layout control with structured masking strategy}: To equip the AR model with more effective layout control, a specially designed structured masking strategy is applied to attention computation. (3) \textbf{Layout GRPO}: To alleviate exposure bias and further improve the generation quality and layout accuracy, a GRPO-based post-training stage is adapted to the next-set-based AR model with an additional layout reward.

\begin{figure}[t]
    \centering
    \includegraphics[width=1\linewidth]{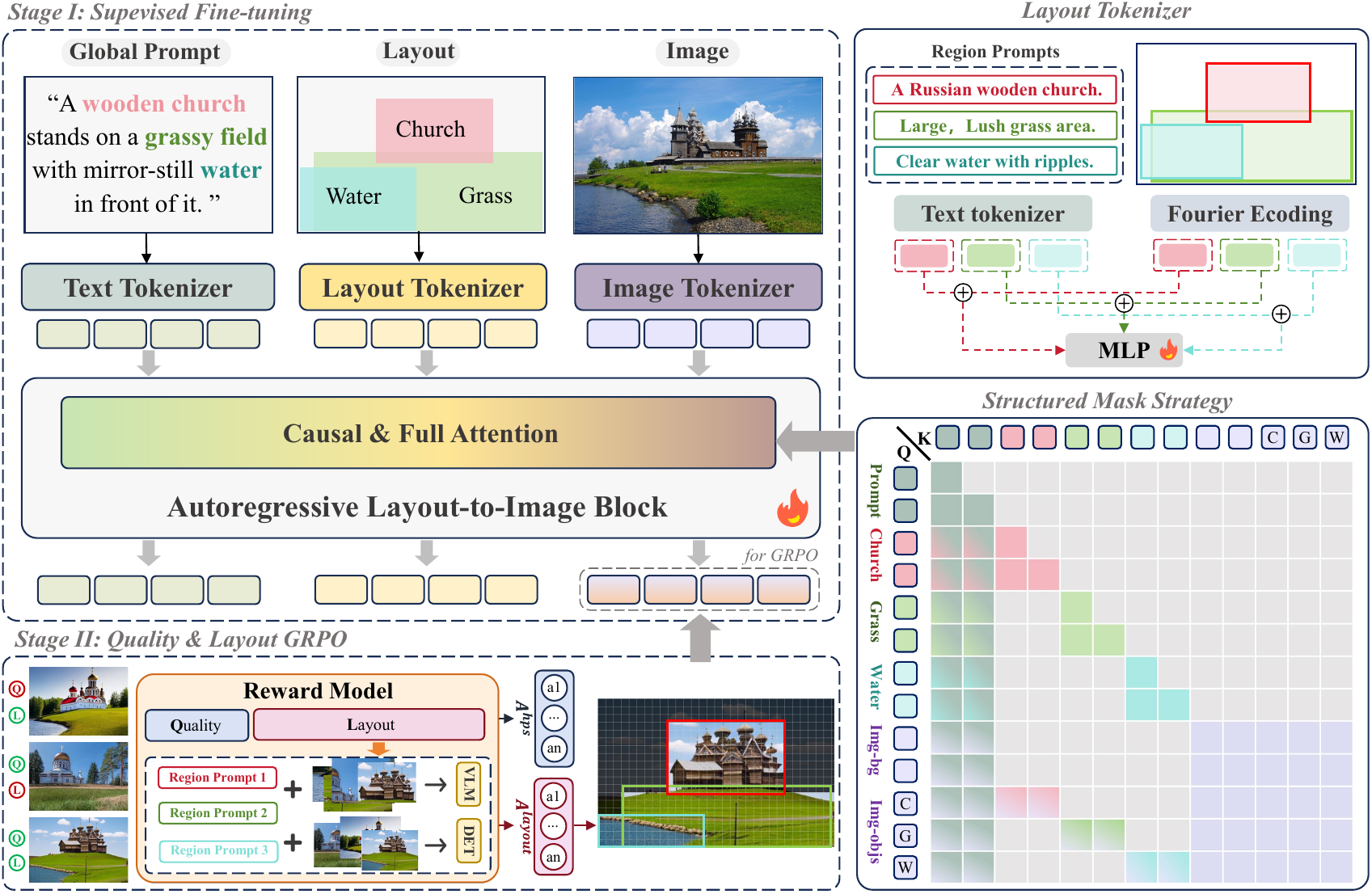}
    \caption{An overview of the proposed framework. In stage 1, the global prompt and image are tokenized using Show-o's original setup. Layout tokens are derived from the layout tokenizer based on the bounding boxes and region prompts. A unified input sequence, formed by the three types of tokens, is fed into the transformer equipped with a dedicated structured masking strategy to enable more effective layout control. After SFT, we further introduce a GRPO-based post-training stage to improve the layout controllability and image quality simultaneously with a dedicated layout reward and image quality reward.}
    \label{fig:pipeline}
\end{figure}

\subsection{Preliminary}
\label{sec:preliminary}

\subsubsection{Next-set-based AR models} 
Traditional AR models generate images token-by-token in raster order, leading to inefficiency for high-resolution images due to long sequences. Next-set-based AR models address this by predicting a set of image tokens at each timestep, reformulating image generation as masked token prediction. Specifically, during inference, all image tokens are initially masked, and at each time step, the model predicts a subset of tokens in parallel. Masked tokens are progressively revealed following a scheduling strategy that reduces the masking ratio from 1.0 to 0. 

\subsubsection{GRPO}
We adopt GRPO algorithm~\cite{shao2024deepseekmath} for post-training, which extends PPO (Proximal Policy Optimization~\cite{schulman2017proximal}) by removing the value function and estimating advantages in a group-relative manner.
The objective is defined as:
\begin{equation}
\label{eq:grpo_obj}
\resizebox{0.8\linewidth}{!}{
$
\begin{aligned}
\mathcal{J}_{GRPO}(\theta) &= \mathbb{E}_{c \sim \mathcal{D}, \{o_i\}_{i=1}^G \sim \pi_{\theta_{\text{old}}}(\cdot|q)}  \\
&\Biggl[ \frac{1}{G} \sum_{i=1}^G \frac{1}{|o_i|}\sum_{t=1}^{|o_i|} \left\{ \min \bigl(r_{i,t} A_{i,t},\, \hat{r}_{i,t} A_{i,t}\bigr) - \beta\, \mathrm{D}_{\mathrm{KL}} \right\} \Biggr]
\end{aligned}
$
}
\end{equation}

GRPO-based post-training involves a trainable policy model $\pi_\theta$, a behavior policy model $\pi_{\theta_{old}}$, and a frozen reference model $\pi_{ref}$, all initialized from a checkpoint obtained through prior supervised fine-tuning (SFT). Given condition $c$ from training data $\mathcal{D}$, the behavior policy model $\pi_{\theta_{old}}$ samples a group of outputs $o_1, o_2, ..., o_G,$ and then optimizes the objective in \cref{eq:grpo_obj} to update $\pi_\theta$. The \( r_{i,t} = \frac{\pi_\theta(o_{i,t} \mid c,o_{i,<t})}{\pi_{\theta_\text{old}}(o_{i,t} \mid c,o_{i,<t})} \) denotes the importance sampling ratio between the trainable policy model \( \pi_\theta \) and behavior policy model \( \pi_{\theta_\text{old}} \), while \( \hat{r}_{i,t} = \mathrm{clip}(r_{i,t}, 1 - \epsilon, 1 + \epsilon) \) applies a clipping threshold \( \epsilon \) to stabilize training. \( A_{i,t} \) represents the advantage of the $t$-th token of $o_i$  obtained by group-wise normalized rewards, since action for the traditional AR model is every single token. The KL divergence \( \mathrm{D}_{\mathrm{KL}} = \mathrm{D}_{\mathrm{KL}}(\pi_\theta \,\|\, \pi_{\text{ref}}) \), weighted by coefficient \( \beta \), regularizes the updated policy towards a fixed reference policy $\pi_{ref}$ to reduce overfitting.

\subsection{Tokenization}

For the prompt and image, we adopt Show-o's setup: the prompt is tokenized by Show-o's text tokenizer, and the image is tokenized into discrete tokens by MAGVITv2~\cite{yu2023magvitv2}. Layout condition consists of multiple objects, where each object is defined by a bounding box $\text{obj}_i^{\text{box}} = [x_0, y_0, x_1, y_1]$ and a textual description $\text{obj}_i^{\text{text}}$. The box specifies the normalized top-left and bottom-right coordinates, while the text provides the semantic description.

To integrate geometric and linguistic information, we adopt a unified layout encoding strategy that embeds spatial boxes and their textual descriptions together. Specifically, we directly use Show-o's text tokenizer $E$ and codebook $C$ to encode the region description of the layout; each bounding box is represented as a set of Fourier embeddings. 
Unlike diffusion models that concatenate text tokens with Fourier embeddings along the channel dimension, we aim to fully leverage the text representations learned during Show-o pretraining. To this end, we concatenate the text tokens and box tokens along the sequence dimension and process them with a zero-initialized MLP equipped with a residual connection, as illustrated in the Layout Tokenizer of \cref{fig:pipeline}.
\begin{gather}
x_i = \mathrm{Concat}\left( C(E(obj_i^{\text{text}})),\; \mathrm{Fourier}(obj_i^{\text{box}}) \right) \label{eq:xi}\\ 
obj_i^{\text{layout}} = x_i + \mathrm{MLP}(x_i) \label{eq:layout_token}
\end{gather}

The layout tokens of object $i$ can be expressed as ~\cref{eq:xi,eq:layout_token}, where $Concat$ denotes concatenation in the sequence dimension. Finally, the input to the AR transformer is a sequence composed of the global text prompt, layout, and image tokens, where the image tokens are partially replaced by [MASK] tokens for SFT training.

\subsection{Structured Masking Strategy}
As shown in \cref{fig:pipeline}, given a sequence of tokens for the prompt, layout, and image tokens in order, a specially designed structured masking strategy is applied to attention computation in transformer blocks as follows.

As for the global text prompt tokens, we adopt the standard causal masking strategy. As for the layout tokens, our attention masking strategy follows three design principles: (1) \textbf{Global Context Awareness}, each layout token is allowed to attend to the global text prompt tokens, enabling it to capture high-level contextual cues and better understand how a specific region relates to the entire image. (2) \textbf{Intra-object Coherence}, layout tokens that belong to the same object are connected under a standard causal mask, allowing them to accumulate fine-grained spatial and semantic information. (3) \textbf{Inter-object Isolation}, for layout tokens belonging to different objects, the masks are designed to isolate them from each other to avoid information interference and preserve object-specific representations.

For image tokens, they attend to all other image tokens and the global prompt tokens, but only to the subset of layout tokens they are associated with, following the principle of \textbf{Region-wise Separation.} If an image token lies within multiple overlapping object boxes, it attends to the associated layout tokens to capture composite layout semantics. In such cases, the global text prompt provides relational context between the overlapping objects, helping the model interpret their semantic relationship and resolve the resulting ambiguity. In this way, the generation process of image tokens is steered by global semantic guidance and associated layout conditions, resulting in precise spatial and semantic alignment while mitigating interference from irrelevant tokens.

Together, these strategies introduce a structured inductive prior that aligns with the sequential organization of prompt, layout, and image tokens, facilitating efficient training without introducing notable architectural modifications or significant parameter overhead.

\subsection{Layout-GRPO}

To further improve image quality and layout controllability, we adopt a GRPO-based post-training framework. However, adapting GRPO to next-set-based AR models for layout-conditioned image generation poses unique challenges. (1) How to define the \textit{action} for next-set-based AR models in GRPO, (2) how to enhance image fidelity without compromising, and potentially improving, the layout controllability of the SFT model.

As introduced in \cref{sec:preliminary}, Show-o follows a \emph{next-set prediction} paradigm. 
At each time step, the model attends to the entire image token sequence and retains only a subset of tokens with high confidence as predictions, while the remaining tokens are masked again,
so the retained token set at each step should be regarded as part of the action. In standard AR models following the next-token-predication paradigm, the context at each time step is simply the generated token prefix, since all prefixes are nested subsequences of the same full sequence, a single forward pass with a causal attention mask can simultaneously construct all step-wise contexts, so the log-probabilities for all time steps can be computed in parallel within one forward evaluation.
In contrast, Show-o uses the full-length input token sequence as context at every step, and the contexts may differ in the token values at the same spatial positions.
Such step-dependent and full-length contexts cannot be reproduced by a fixed attention mask, making it impractical to compute all step-wise log-probabilities in a single forward pass. To address this issue, during rollouts with $\pi_{\theta_{\text{old}}}$, we explicitly record the retained token set together with their spatial positions at each time step as a composite action, and treat the current input token sequence as the corresponding state.
The $\pi_\theta$ and $\pi_{\theta_{\text{ref}}}$ then replay the same state--action trajectory step by step, allowing us to compute $\pi_\theta(o_{i,t} \mid c, o_{i,<t})$ and $\pi_{\theta_{\text{ref}}}(o_{i,t} \mid c, o_{i,<t})$ under identical contexts.

To preserve layout controllability while improving the image quality, we propose a dedicated layout reward that integrates well with the image quality reward (HPS v2.1 score~\cite{wu2023hps}), leading to improved overall performance. Specifically, for each bounding box in the layout, we evaluate the corresponding image region with its detailed textual description in a visual question answering (VQA) manner using a vision-language model~\cite{Qwen3-VL} (VLM), obtaining a binary \textbf{VQA score} $S_{\text{VQA}}$ that reflects semantic and attribute consistency. However, a positive VQA score is given as long as the object’s main part appears in the specified region, regardless of precise spatial boundaries, and it does not penalize false positives when the object appears elsewhere. To address this, we further introduce the mIoU score and the precision score to encourage more accurate locations and avoid false positives. Specifically, we employ Grounding DINO~\cite{liu2024g-dino} to detect the generated image according to the given category labels, and compute the mIoU and precision by comparing the detected results with the layout conditions. The weighted sum of these score forms the layout reward $R_{\text{layout}}$, as shown in \cref{eq:layout_reward}.
\begin{equation}
\label{eq:layout_reward}
R_{\text{layout}} = \lambda_{\text{VQA}} \cdot S_{\text{VQA}} 
+ \lambda_{\text{mIoU}} \cdot S_{\text{mIoU}} 
+ \lambda_{\text{prec}} \cdot S_{\text{prec}},
\end{equation}

\vspace{-3mm}

\begin{equation}
\label{eq:overall advantange}
A_{i,t} = 
\omega^{\text{hps}} \cdot A_i^{\text{hps}} + \omega^{\text{layout}} \cdot A_i^{\text{layout}}
\end{equation}

The overall advantage is defined in \cref{eq:overall advantange} as a weighted combination of the HPS advantage $A_i^{\text{hps}}$ and the layout advantage $A_i^{\text{layout}}$. A larger $\omega^{\text{layout}}$ is further applied to image tokens located within the bounding boxes specified by the layout condition, thereby encouraging the model to focus on layout-aligned regions during policy optimization.

\section{Experiments}
\label{sec:expe}

\subsection{Datasets and Evaluation Metrics}
\textbf{Data Preparation.} For SFT training, we use the LayoutSAM~\cite{zhang2024creatilayout} dataset. For GRPO-based post-training, to ensure reliable layout rewards, we filter out samples from LayoutSAM that contain semantically redundant instances, and we further remove samples whose Grounding-DINO~\cite{liu2024g-dino} detection results fail to reach mIoU and precision above 0.9.

\noindent
\textbf{Benchmark and Evaluation Metric.} 
We evaluate our method across various benchmarks, including the Layout-to-Image (L2I) datasets LayoutSAM-Eval~\cite{zhang2024creatilayout} and OverLayBench~\cite{lioverlaybench}, as well as the compositional Text-to-Image (T2I) benchmark T2I-CompBench~\cite{huang2023t2icomp}. For LayoutSAM-Eval, we assess layout control through spatial positioning and attribute alignment, specifically focusing on color, texture, and shape consistency. Image quality is measured by human preference scores (IR~\cite{xu2023imagereward}, Pick~\cite{kirstain2023pick}), text-image alignment (CLIP~\cite{radford2021clip}), and perceptual quality (FID~\cite{heusel2017fid}). Furthermore, 895 samples are selected from LayoutSAM-Eval using Grounding-DINO to evaluate mIoU and precision. OverLayBench addresses more demanding layout conditions characterized by multiple overlapping bounding boxes with semantically similar categories. On this benchmark, layout control is evaluated via mIoU, O-mIoU (mIoU calculated on overlapped regions), SR$\text{E}$ (Success Rate of Entity), and SR$\text{R}$ (Success Rate of generated spatial Relationships). Semantic consistency is measured at both image and region levels using CLIP$_\text{G}$ and CLIP$_\text{L}$, respectively. Regarding T2I-CompBench, we use metrics that intersect with layout-related capabilities, including color, shape, texture, spatial reasoning, and numeracy.

\subsection{Training Details}

We adopt Show-o as the foundation model, with a training resolution of $512\times512$. For the SFT stage, we use the AdamW~\cite{loshchilov2017decoupled} optimizer with a fixed learning rate of $2\times10^{-5}$, training for 30,000 iterations with a global batch size of 128. Trainable components include the entire AR transformer and the MLP in the layout tokenizer. The SFT stage took 3 days on 8 AMD MI300X GPUs.

In GRPO-based post-training, the number of groups is set to 4, the number of generation steps to 10, the batch size to 28, and the learning rate to $1\times10^{-4}$, with training conducted for 100 iterations. Each transformer block is fine-tuned using LoRA~\cite{hu2022lora} with a rank of 256. For rollouts, the temperature is set to 1 and the classifier-free guidance scale to 5. Following DanceGRPO~\cite{xue2025Dancegrpo}, to mitigate the training instability introduced by rollouts with classifier-free guidance, each policy update relies exclusively on newly generated rollouts, and $\beta$ in \cref{eq:grpo_obj} is set to 0. The weights $\omega^{\text{hps}}$ and $\omega^{\text{layout}}$ are set to 1. When image tokens are located within layout-specified bounding boxes, $\omega^{\text{layout}}$ is increased to 1.2 to reinforce layout control. For the layout reward in~\cref{eq:layout_reward}, $\lambda_{\text{VQA}}$, $\lambda_{\text{mIoU}}$, and $\lambda_{\text{prec}}$ are set to 0.5, 0.5, and 0.1, respectively. The GRPO stage took 4 hours on 8 AMD MI250 GPUs. Compared with the SFT stage, the GRPO stage introduces only \textbf{minimal} additional training overhead, while yielding clear improvements in both image quality and layout controllability.

\vspace{-3mm}
\begin{figure}[h]
    \centering
    \includegraphics[width=1\linewidth]{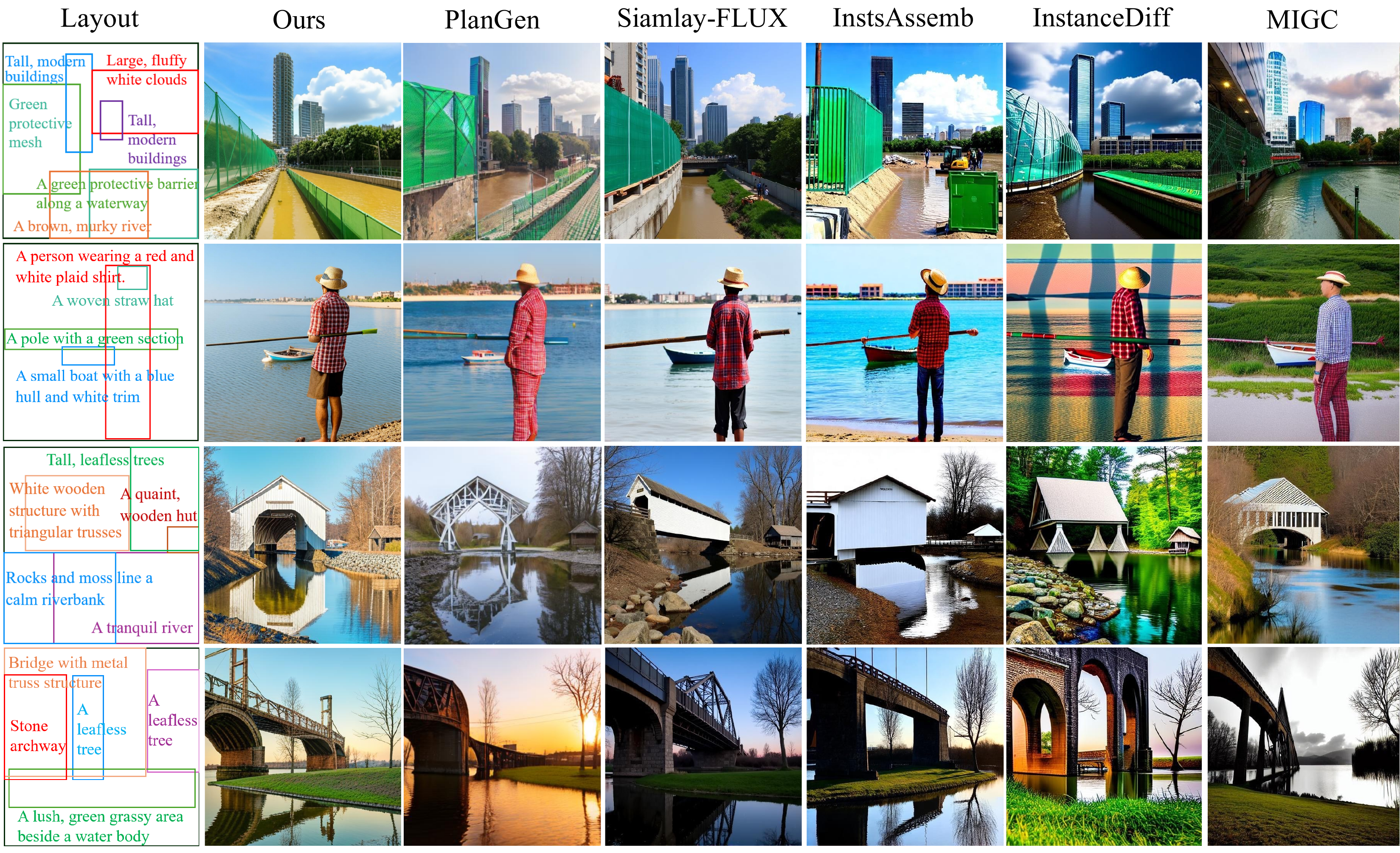}
    \caption{Qualitative results on LayoutSAM-Eval. Under complex layouts, previous methods often fail to generate all specified objects or capture their fine-grained attributes, whereas SMARLI reliably adheres to the layout and generates all required objects.
    }
    \label{fig:compare_layoutsam}
\end{figure}
\vspace{-7mm}

\subsection{Evaluation on Layout-to-Image Generation}

We compare SMARLI with state-of-the-art L2I models on LayoutSAM-Eval and OverLayBench.
For LayoutSAM-Eval, most results are adopted from CreatiLayout~\cite{zhang2024creatilayout}, while the scores of InstsAssemb and PlanGen are taken from their original papers.
For OverLayBench, we use the official benchmark results when available and reproduce the results of InstsAssemb and PlanGen using their released code.
In addition, we validate SMARLI on another autoregressive image generation model, Janus-Pro-1B, which follows standard next-token prediction.
It achieves performance comparable to SMARLI-Show-o.
Detailed results and analysis are provided in the supplementary material.

\vspace{-3mm}

\begin{table}[]
\caption{Quantitative results on LayoutSAM-Eval~\cite{zhang2024creatilayout}. Spa, Col, Tex, Sha, and Pre denote Spatial, Color, Texture, Shape, and Precision, respectively. \textbf{Bold} and \underline{underline} indicate the best and second-best results. * means it is trained on LayoutSAM.} 
\label{tab: results on LayoutSAM-Eval}
\centering
\tabcolsep=0.06cm
\ra{1.1}
\scalebox{0.86}{
\begin{tabular}{@{}lccccccccccc@{}}
\toprule
\multirow{2}{*}{\textbf{Method}} & \multirow{2}{*}{Params} & \multicolumn{6}{c}{Layout Control} & \multicolumn{4}{c}{Image Quality} \\ \cmidrule(l){3-8} \cmidrule(l){9-12} 
                    &  & Spa.$\uparrow$   & Col.$\uparrow$   & Tex.$\uparrow$  & Sha.$\uparrow$  & mIoU$\uparrow$  & Pre.$\uparrow$  &IR$\uparrow$     & Pick$\uparrow$  & CLIP$\uparrow$  & FID$\downarrow$  \\ \midrule
\rowcolor[RGB]{245,245,245} GT Images          &  -  & 98.95	&98.45	&98.90	 &98.80  &98.36 &100.0                & -  &  -  & -  & - \\  \midrule
GLIGEN~\cite{li2023gligen}             & ~1.2B &  77.53     & 49.41   & 55.29    &  52.72 & 69.62 & 70.99  & -10.31   & 20.78  &32.42 & 21.92 \\
InstanceDiff~\cite{wang2024instancediffusion}            & ~1.4B & 87.99     & 69.16   & 72.78   & 71.08 & \underline{81.29} & 72.34  & 9.14   & 21.01  & 31.40  & 19.67 \\
MIGC~\cite{zhou2024migc}                & ~1.2B & 85.66     & 66.97   & 71.24    & 69.06 & 70.99 & 74.10  & -13.72  & 20.71  & 31.36 & 21.19 \\
HiCo~\cite{cheng2024hico}           & ~1.4B & 87.04     & 69.19   & 72.36    & 71.10 & 81.20 & 71.74  & 12.36   & 21.70  & 32.18 & 22.61 \\
SiamLay{-\tiny{SD3}}*~\cite{zhang2024creatilayout}         & ~3B & 92.67    & 74.45   & 77.21    & 75.93 & 68.43 & 63.22  & 69.47   & \underline{22.02}  & \textbf{34.01} & 19.10 \\
InstsAssemb*~\cite{xianginstanceassemble}         & ~2B & 94.97    & 77.53   &  80.72    & 80.11  & 73.33 & 74.19  & 70.14   & 21.75  & 33.08 & 20.24 \\ 
SiamLay{-\tiny{Flux}}*~\cite{zhang2024creatilayout}         & 22B & \textbf{95.67}   & 80.71   &  83.53    & 82.80 & 78.01 & 78.33  & \textbf{80.48}   & \textbf{22.16}  & \underline{33.92} & \underline{16.12} \\
\midrule
PlanGen*~\cite{he2025plangen}         & 1.5B & 92.21   & \underline{82.69}   &  \underline{86.53}    & \underline{85.36} & 78.04 & \underline{81.15}  & 39.58   & 21.43  & 31.96 & \textbf{13.91} \\
\midrule
\textbf{SMARLI}*         & 1.3B & \underline{95.55}    & \textbf{87.35}   & \textbf{90.90}    & \textbf{89.82} & \textbf{81.31} & \textbf{81.84}  & \underline{74.81}   & 21.81  & 33.14 & 18.16 \\ 
\bottomrule
\end{tabular}}
\end{table}

\vspace{-10mm}

\begin{table}[]
  \centering
  \caption{Quantitative results on OverLayBench-Complex. }
  \label{tab:results_on_overlay}
  \setlength{\tabcolsep}{2pt}
  \setlength\dashlinedash{0.5pt}
  \setlength\dashlinegap{2pt}
  \setlength\arrayrulewidth{0.3pt}
  
  \renewcommand{\arraystretch}{1.1}
  \scalebox{0.98}{
\begin{tabular}{@{}lccccccc@{}}
\toprule
\textbf{Method} & Params & mIoU $\uparrow$ & O-mIoU $\uparrow$ & SR$_\text{E}$ $\uparrow$ & SR$_\text{R}$ $\uparrow$ & CLIP$_\text{G}$ $\uparrow$ & CLIP$_\text{L}$ $\uparrow$  \\

\midrule
GLIGEN~\cite{li2023gligen} & 1.2B & 50.79 & 23.85  & 41.70 & 79.93 & 33.92 & 22.75 \\
InstanceDiff~\cite{wang2024instancediffusion} & 1.4B & 53.68 &25.63 & 66.02 & 80.34 & 32.33 & 25.53  \\
MIGC~\cite{zhou2024migc} & 1.2B & 40.04 & 13.26 & 47.80 & 74.48 & 31.93 & 24.20  \\
HiCo~\cite{cheng2024hico} & 1.4B & 46.56 & 20.35 & 48.88 & 75.19 & 33.15 & 24.41 \\
SiamLay{-\tiny{SD3}}~\cite{zhang2024creatilayout} & 3B & 44.24 & 18.05 & 52.10 & 79.98& 36.55 & 24.76  \\
SiamLay{-\tiny{FLUX}}~\cite{zhang2024creatilayout} & 22B & \underline{54.50} &	\underline{28.97}&	69.72&	\underline{86.45}&	\textbf{36.72}	& 24.85 \\
EliGen~\cite{zhang2025eligen} & 12B & 52.53 & 26.19 & \underline{74.03} & 84.09 & 36.18 & \underline{25.92}  \\
InstsAssemb~\cite{xianginstanceassemble} & 2B & 49.12 & 23.26 & 58.55 & 83.22 & 35.29 & 24.32 \\
\midrule
PlanGen~\cite{he2025plangen} & 1.5B & 51.21 & 23.53 & 67.79 & 85.36 & 35.22 & 19.17 \\
\midrule
SMARLI & 1.3B & \textbf{56.42} & \textbf{29.59} & \textbf{80.52} & \textbf{87.81} & \underline{36.57} & \textbf{26.48}\\
\bottomrule
\end{tabular}
}
\end{table}

\vspace{-3mm}

The quantitative results are presented in \cref{tab: results on LayoutSAM-Eval} and \cref{tab:results_on_overlay}. On LayoutSAM-Eval, the SMARLI demonstrates strong layout control performance, excelling in spatial, color, texture, and shape control. It also achieves promising quality performance, with competitive results in these image quality metrics. On OverlayBench-Complex, SMARLI maintains more robust layout alignment under layouts with multiple overlapping bounding boxes. For completeness, results on the Simple and Regular subsets are provided in the supplementary material, where SMARLI consistently outperforms other methods, further demonstrating its effectiveness across varying layout complexities.
\vspace{-3mm}
\begin{figure}[]
    \centering
    \includegraphics[width=1\linewidth]{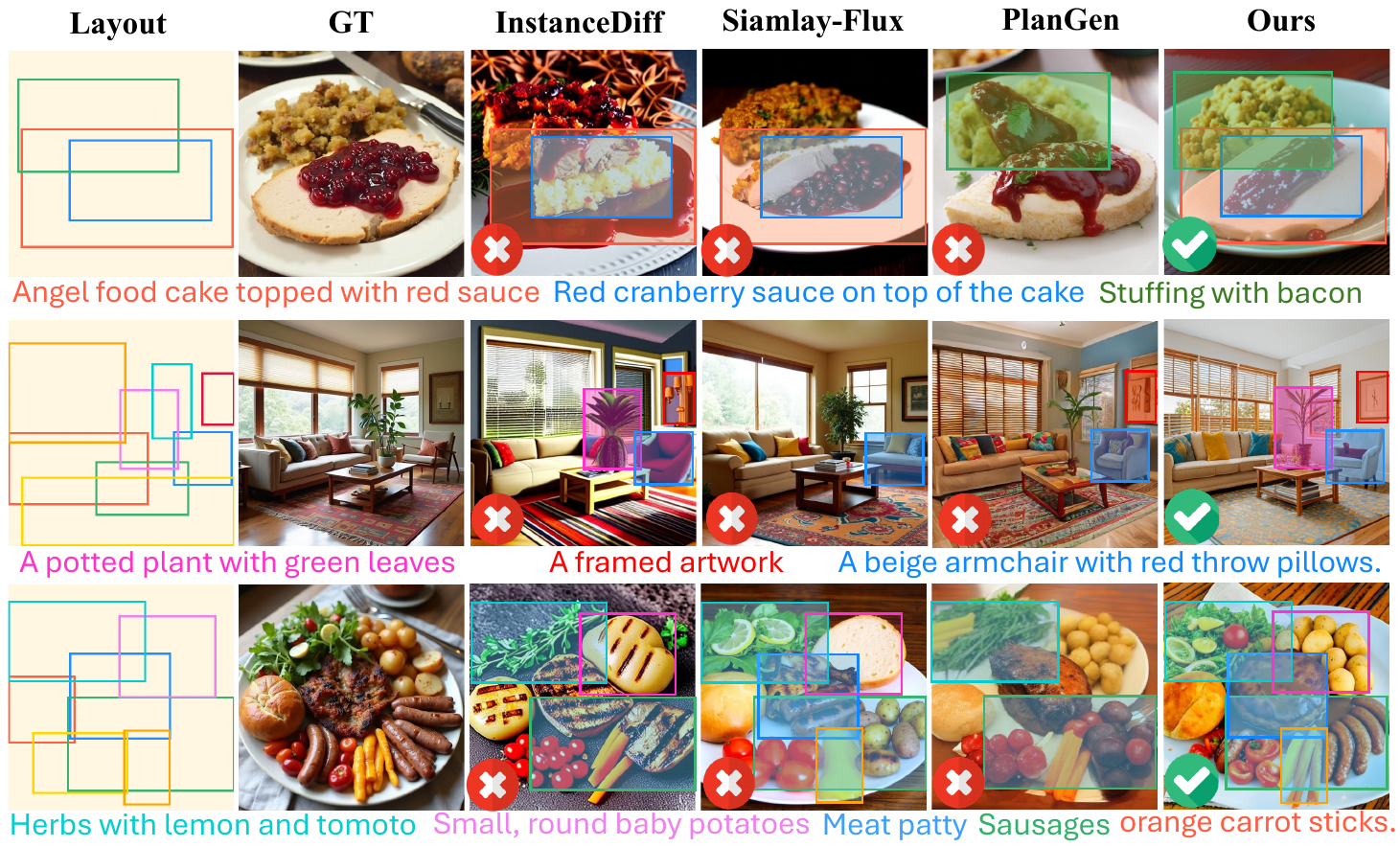}
    \caption{Qualitative results on OverLayBench.
    }
    \label{fig:compare_overlay}
\end{figure}

% \vspace{-3mm}

The qualitative results are shown in \cref{fig:compare_layoutsam} and~\cref{fig:compare_overlay}. Images generated by our method demonstrate superior layout control, exhibiting more accurate spatial positioning and faithful attribute rendering than existing approaches. These results highlight the effectiveness of the proposed masking strategy, which enhances the semantic representation of layout tokens and mitigates interference between tokens from different objects, together with the benefits introduced by the well‑designed layout reward in GRPO‑based post‑training. The compelling quantitative and qualitative results demonstrate the potential of AR models as strong competitors to diffusion models.

\subsection{Evaluation on Text-to-Image Generation}

% 用vspace调出来的成功版
\begin{wraptable}{r}{0.60\textwidth} % {r} 表示靠右浮动，0.55\textwidth 是表格区域宽度
\vspace{-3.5em}
\centering
\caption{Quantitative results on T2I-CompBench.}
\label{tab:t2i_comparison}
\scalebox{0.7}{
\setlength{\tabcolsep}{2pt} % 适当放宽列间距，替代原本过紧的 1pt
\renewcommand{\arraystretch}{1.2} 
\begin{tabular}{lccccc}
\toprule
Model & Spatial $\uparrow$ & Color $\uparrow$ & Shape $\uparrow$ & Texture $\uparrow$ & Numeracy $\uparrow$ \\
\midrule
% Attn-Exct~\cite{chefer2023attend}   & 14.6 & 64.0 & 45.2 & 59.6 & 47.7 \\
SDXL~\cite{podell2023sdxl}       & 21.3 & 58.8 & 46.9 & 53.0 & 49.9 \\
PixArt-$\alpha$~\cite{chen2023pixart} & 20.6 & 66.9 & 49.3 & 64.8 & 50.6 \\
DALLE-3~\cite{betker2023improving}     & 28.7 & \underline{77.9} & \textbf{62.1} & \underline{70.4} & 58.8 \\
% Janus-Pro-7B~\cite{chen2025janus} & 20.6 & 63.6 & 35.3 & 49.4 & xxx \\
% Show-o~\cite{xie2024showo}& 20.4 &  56.2 & 41.0 & 46.8 & 51.8 \\
\midrule
Show-o~\cite{xie2024showo} & \underline{39.6} & 73.0 & 52.4 & 67.7 & \underline{62.3} \\
\midrule
SMARLI & \textbf{49.7} & \textbf{79.6} & \underline{59.2} & \textbf{72.3} & \textbf{68.4} \\
\bottomrule
\end{tabular}}
\vspace{-2em} % 微调表格下方的空白，避免环绕文字过远
\end{wraptable}

Following Creatilayout~\cite{zhang2024creatilayout}, we conduct experiments on T2I-CompBench to evaluate the benefit brought by the additional layout condition. Since the T2I-CompBench only provides the text prompts, we use GPT-5.2 to generate the corresponding layouts. As shown in \cref{tab:t2i_comparison}, conditioned on the text prompt and layouts, SMARLI achieves a significant improvement in spatial adherence (from 39.6 to 49.7). Additionally, the benefits of the layout are also reflected in the consistent improvements in numeracy and attribute binding.

\subsection{Ablation Study}

In this section, we conduct ablation studies on the structure masking strategy and GRPO-based Post-training. Further ablation studies on the layout tokenizer and $\omega^{\text{layout}}$ can be found in the supplementary material.
\subsubsection{Masking Strategy.}

\begin{figure}[htbp]
    \centering
    \begin{minipage}[t]{0.44\linewidth}
        \vspace{0pt} 
        \centering
        
        \setlength{\belowcaptionskip}{10pt} 
        
        \captionof{table}{\small Ablation study on different masking strategies. Compared to the baseline, a noticeable performance degradation in spatial positioning and attribute binding is observed when the LP or Local Causal is removed.}
        \label{tab:quantitative_results_on_mask}
        
        \renewcommand{\arraystretch}{1.2} 
        
        \resizebox{\linewidth}{!}{
            \begin{tabular}{@{}lcccccc@{}}
            \toprule                
            \textbf{Mask Strategy} & Spa. & Col. & Tex. & Sha. & mIoU & Pre.\\
            \midrule
            SMARLI-SFT & \textbf{95.26} & \textbf{86.17} & \textbf{89.79} & \textbf{88.56} & \underline{80.08} & \textbf{80.69} \\
            w/o LP     & \underline{95.17} & 79.83 & 84.35 & 82.74 & \textbf{80.13} & \underline{80.55} \\
            w/o Local Causal & 93.93 & \underline{83.22} & \underline{87.12} & \underline{85.68} & 78.39 & 77.04 \\
            \bottomrule
            \end{tabular}
        }
    \end{minipage}% <-- 
    \hfill 
    \begin{minipage}[t]{0.54\linewidth}
        \vspace{0pt} 
        \centering
        \includegraphics[width=\linewidth, height=4.5cm, keepaspectratio]{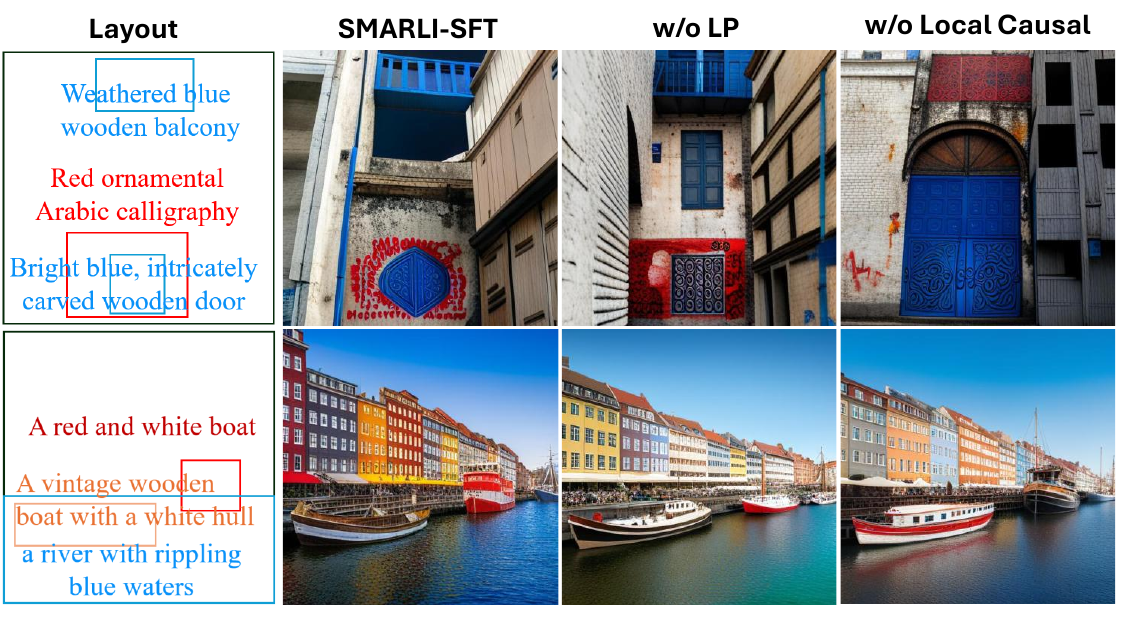}
        
        \setlength{\abovecaptionskip}{2pt} 
        
        \caption{\small Visualization of ablation study on masking strategies.}
        \label{fig:viz_mask_ablation}
    \end{minipage}
\end{figure}

We first conduct an ablation study on the effectiveness of the proposed masking strategy. The model without GRPO-based post‑training (SMARLI‑SFT) is used as the baseline, and the results are reported in \cref{tab:quantitative_results_on_mask}. We examine two variant masking strategies: (1) layout tokens are isolated from global prompt tokens (w/o LP); (2) layout tokens are no longer treated as independent units per object but are instead processed as a whole (w/o Local Causal).

\vspace{-3mm}
\vspace{-1mm}

% \textbf{Analysis:} 
For the w/o LP, layout tokens are prohibited from attending to global prompt tokens. As a result, layout tokens cannot leverage the contextual semantic information provided by global prompt tokens, significantly undermining the model’s ability to control object attributes. As illustrated in \cref{fig:viz_mask_ablation}, it causes failures in attribute rendering, highlighting the critical role of Global Context Awareness.

For w/o Local Causal, layout tokens attend to both global prompt tokens and preceding layout tokens, while image tokens are not restricted to their corresponding regions. This indiscriminate mixing of attribute and spatial information across objects degrades performance, especially in spatial metrics. As shown in \cref{fig:viz_mask_ablation}, it causes misaligned object spatial positioning and attribute confusion, underscoring the importance of Intra-object Coherence, Inter-object Isolation, and Region-wise Separation.

\subsubsection{GRPO-based Post-training.}

\begin{table}[h]
\caption{Ablation study on GRPO-based post-training.}
\label{tab:ablation on GRPO}
\setlength{\tabcolsep}{2mm}
\centering
\small
\scalebox{0.95}{
\begin{tabular}{ccc|cccccc|cc}
\toprule
\multirow{2}{*}{\raisebox{-5.7ex}{\rotatebox{90}{\textbf{HPS}}}} &
\multirow{2}{*}{\raisebox{-5.7ex}{\rotatebox{90}{\textbf{VQA}}}} &
\multirow{2}{*}{\raisebox{-5.7ex}{\rotatebox{90}{\textbf{DET}}}} &
\multicolumn{6}{c|}{\textbf{Layout Control}} & 
\multicolumn{2}{c}{\textbf{Image Quality}} \\
\cmidrule(lr){4-9} \cmidrule(lr){10-11}
& & & \rotatebox{90}{Spatial} & \rotatebox{90}{Color} & \rotatebox{90}{Texture} & \rotatebox{90}{Shape}  & \rotatebox{90}{mIoU} & \rotatebox{90}{Pre} 
  & \rotatebox{90}{IR} & \rotatebox{90}{Pick} \\
\midrule
\gxmark & \gxmark & \gxmark &  95.26 & 86.17 & 89.79 & 88.56 & 80.08 & 80.69 & 60.88 & 21.45 \\ % SFT
\midrule
\cmark & \gxmark & \gxmark & 93.82 & 83.96 & 87.72 & 86.78 & 77.79 & 77.89 & \textbf{76.62} & \textbf{21.96} \\ % hps only
\gxmark & \cmark & \cmark &  95.41 & 87.04 & \underline{90.15} & \underline{89.32} & \underline{80.76} & 80.81 & 58.34 & 21.28 \\ % layout only
% \cdashline{1-11}
\cmark & \cmark & \gxmark & \underline{95.48} & \underline{87.30} & 90.01 & 88.97 & 77.73 & 74.50 & 73.36 & 21.69 \\ % hps + layout vqa only
\cmark & \gxmark & \cmark &  95.13 & 85.61 & 89.45 & 88.29 & 80.70 & \underline{80.88} & 73.05 & 21.53 \\ % hps + layout det only
\midrule
\cmark & \cmark & \cmark & \textbf{95.55}  & \textbf{87.35} & \textbf{90.90 } & \textbf{89.82 } & \textbf{81.31} &\textbf{ 81.84} & \underline{74.81 } & \underline{21.81} \\ % full
\bottomrule
\end{tabular}}

\end{table}

\begin{figure}[th]
  \centering
  \begin{subfigure}{0.49\linewidth}
    \includegraphics[width=1\linewidth]{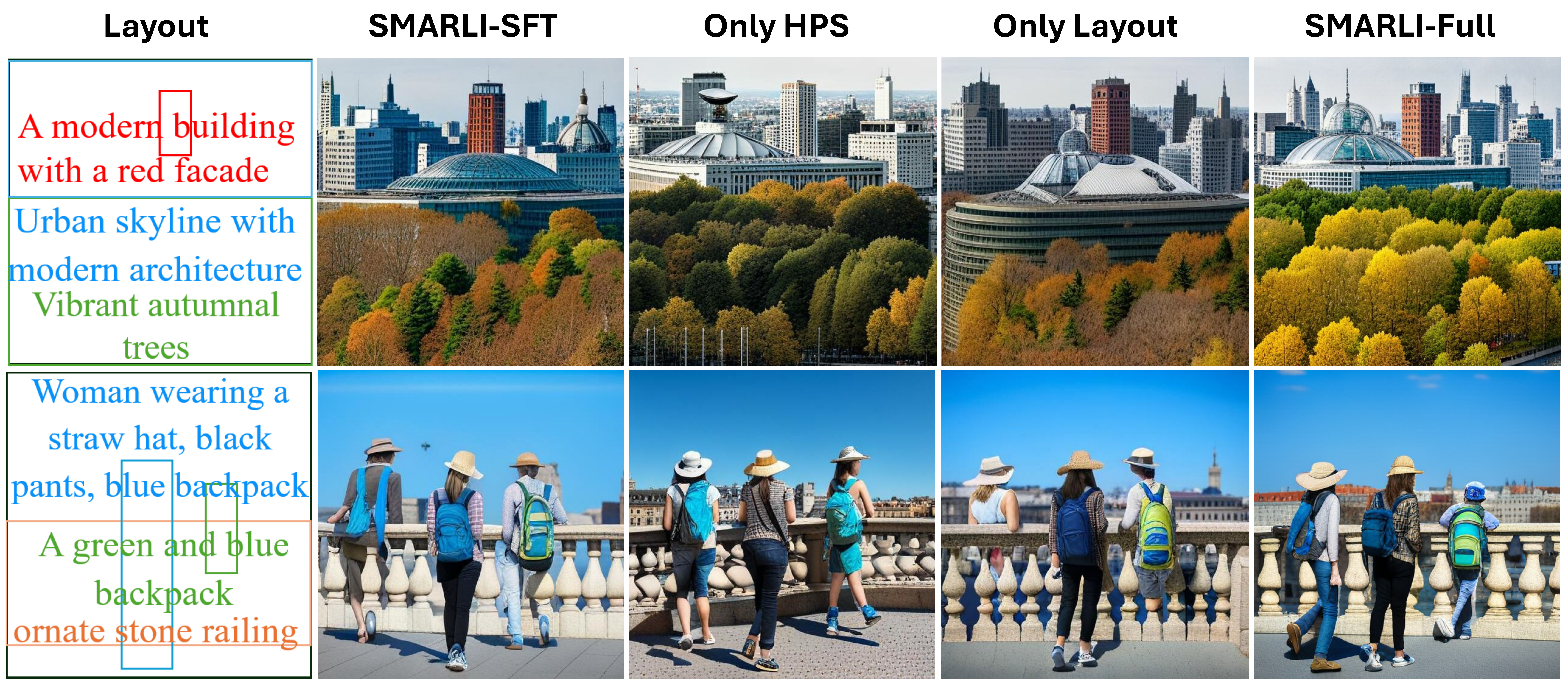}
    \caption{Ablation study on hps and layout rewards.}
    \label{fig:viz_reward_ablation}
  \end{subfigure}
  \hfill
  \begin{subfigure}{0.5\linewidth}
    \includegraphics[width=1\linewidth]{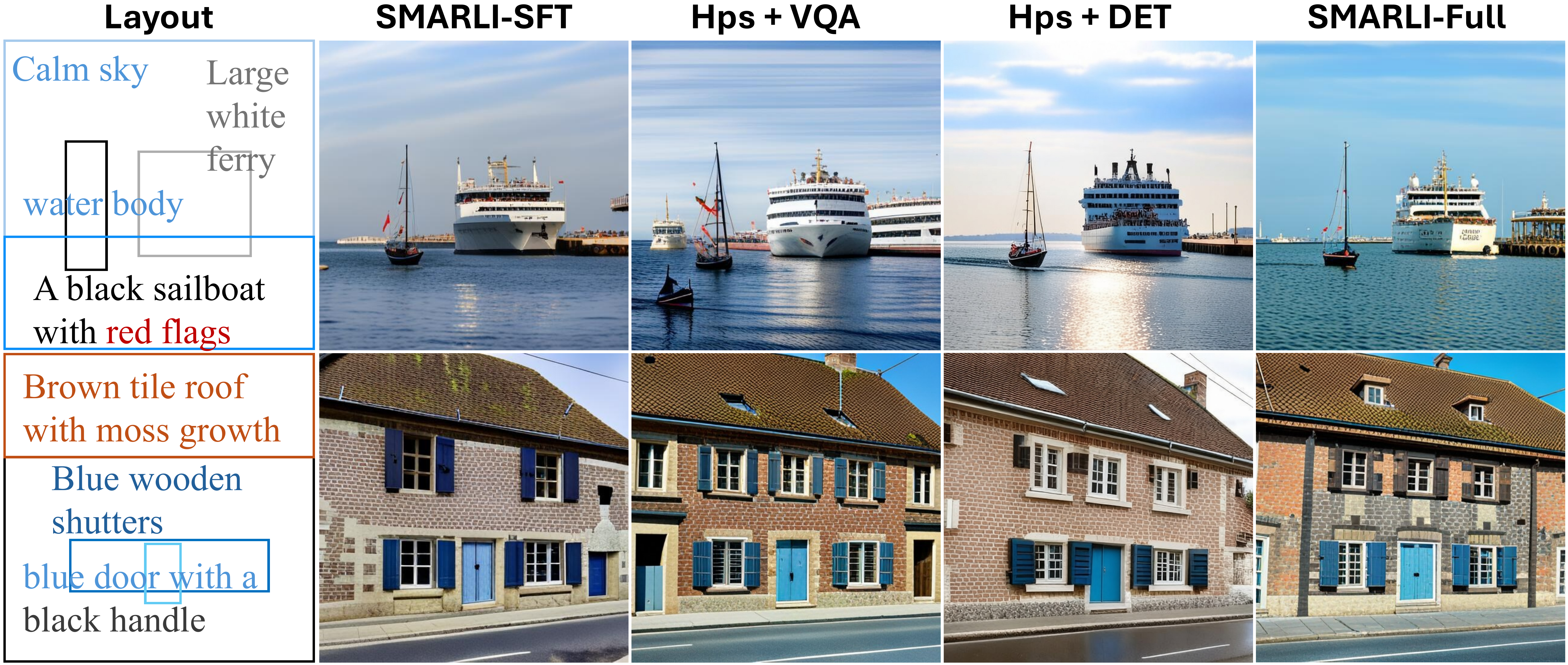}
    \caption{Detailed ablation study on layout reward.}
    \label{fig:viz_layout_reward_ablation}
  \end{subfigure}
  \caption{Visualization of ablation study on rewards.}
  \label{fig:short}
\end{figure}

We conduct ablation studies starting from an SFT baseline to evaluate GRPO-based post-training. As shown in \cref{tab:ablation on GRPO} and \cref{fig:viz_reward_ablation}, using HPS alone (``Only HPS'') improves image quality but reduces layout consistency, particularly in color accuracy. Using layout reward only (``Only Layout'') enhances layout alignment but degrades the image quality. Combining both rewards (``SMARLI-Full'') achieves improved visual quality and layout adherence, with HPS further reinforcing global prompt-image consistency. 

% \vspace{-3mm}
% \vspace{-3mm}

Moreover, we ablate the layout reward design. We divide the layout reward into the VQA score and the detection (DET) score (mIoU score and precision score). As shown in \cref{tab:ablation on GRPO} and ~\cref{fig:viz_layout_reward_ablation}, together with hps score, using only the VQA score (``Hps$+$VQA'') improves spatial positioning and attribute alignment but neglects low-IoU and false-positive issues. Conversely, using only the detection score (``Hps$+$DET'') effectively reduces low-IoU and false positives, yet it compromises semantic consistency with the detailed textual description. The combination of both rewards leads to more balanced and superior performance.

\section{Conclusion}
\label{sec:conclusion}
In this paper, we have presented SMARLI, a novel framework that effectively integrates layout constraints into AR-based T2I generation. Our unified input design and structured masking strategy enable precise spatial control and coherent attribute rendering, while GRPO-based post-training with image-quality and layout rewards further improves generation fidelity and layout alignment. Experimental results demonstrate that SMARLI achieves superior layout control while maintaining the simplicity of AR models. This work provides valuable insights for incorporating layout conditions into AR models, advancing the field of AR-based controllable image generation.
\section{Acknowledgment}
\label{sec:ack}
This work was supported by the National Natural Science Foundation of China under Grant No. 62441231, 62472065, U23B2010,  Talent Fund of Liaoning Province under Grant No. XLYC2503161, and the Open Research Fund from Guangdong Laboratory of Artificial Intelligence and  Digital Economy (SZ), under Grant No. GML-KF-26-08.

\bibliographystyle{splncs04}
\bibliography{main}
\end{document}